\documentclass[runningheads]{llncs}
\usepackage[T1]{fontenc}
\usepackage{graphicx}
\usepackage{booktabs}
\usepackage{algorithm}
\usepackage{algpseudocode}
\usepackage{amsmath}
\usepackage[table,xcdraw]{xcolor}
\usepackage{comment}
\usepackage{multirow}

\begin{document}

\title{ChannelDropBack: Forward-Consistent Stochastic Regularization for Deep Networks}

\titlerunning{ChannelDropBack}

\author{Evgeny Hershkovitch Neiterman\inst{1} \and
Gil Ben-Artzi\inst{1}}
\authorrunning{E. H. Neiterman et al.}
%
\institute{School of Computer Science, Ariel University, Ariel, Israel \and
\email{\{neiterman,gilba\}@ariel.ac.il}} 

\maketitle

\begin{abstract}
Incorporating stochasticity into the training process of deep convolutional networks is a widely used technique to reduce overfitting and improve regularization. Existing techniques often require modifying the architecture of the network by adding specialized layers, are effective only to specific network topologies or types of layers - linear or convolutional, and result in a trained model that is different from the deployed one. We present ChannelDropBack, a simple stochastic regularization approach that introduces randomness only into the backward information flow, leaving the forward pass intact. ChannelDropBack randomly selects a subset of channels within the network during the backpropagation step and applies weight updates only to them. As a consequence, it allows for seamless integration into the training process of any model and layers without the need to change its architecture, making it applicable to various network topologies, and the exact same network is deployed during training and inference. Experimental evaluations validate the effectiveness of our approach, demonstrating improved accuracy on popular datasets and models, including ImageNet and ViT. Code is available at \url{https://github.com/neiterman21/ChannelDropBack.git}.
  \keywords{Regularization \and Stochastic Training \and Deep Learning}
\end{abstract}

\section{Introduction}
\label{sec:intro}

Deep neural networks have achieved remarkable success in fields such as computer vision and natural language processing. Numerous methods have been implemented to enhance their training~\cite{ioffe2015batch,kingma2014adam} and runtime performance~\cite{menghani2023,ofir2022smm,10095537}. However, as these networks become deeper and more complex, overfitting continues to be a challenge~\cite{9380520}. To address this, researchers have developed various regularization techniques. One widely used approach to mitigate overfitting and improve regularization involves incorporating stochasticity.

Existing stochastic regularization approaches, such as Dropout \cite{srivastava2014dropout} and DropConnect \cite{wan2013regularization}, have shown promise in reducing overfitting. Dropout works by randomly deactivating a subset of neurons during training, thereby preventing complex co-adaptations among neurons. DropConnect, on the other hand, generalizes this idea by randomly removing connections between neurons instead of deactivating the neurons themselves. While effective, these methods require architectural modifications and introduce a discrepancy between the trained and deployed models, as only a subset of the connections or neurons are used during training, while all connections or neurons participate at inference.

Other approaches like Stochastic Depth \cite{huang2016deep} and DropPath \cite{larsson2016fractalnet} address the training of very deep networks by randomly dropping entire layers or paths during training. Despite their effectiveness, these methods are primarily applicable to specific network topologies such as residual networks. Techniques like DropBlock \cite{ghiasi2018dropblock} and Spatial Dropout \cite{tompson2015efficient} introduce structured stochasticity by dropping contiguous regions of feature maps or channels, but they still require specialized layers or are limited to particular network structures.

In this paper, we present ChannelDropBack, a stochastic regularization technique that addresses the limitations of existing methods. The primary objective of ChannelDropBack is to bridge the gap between training and inference phases while enhancing the model's generalization capabilities. Our method ensures that the same network structure is used during both phases, thus eliminating any potential discrepancies that might arise due to differences between training and deployment. It introduces randomness exclusively in the backward pass of the neural network, leaving the forward pass intact, by randomly selecting a subset of layers and channels within the network during backpropagation and applies weight updates only to these selected components. 

We conduct extensive experiments on popular datasets such as ImageNet, CIFAR-100, and CIFAR-10, using a variety of architectures including ResNet, EfficientNet, and ViT-B. Our results demonstrate that ChannelDropBack consistently improves accuracy and robustness compared to traditional training methods and other stochastic regularization techniques. The experiments highlight the versatility and efficacy of ChannelDropBack across different datasets and network architectures, showcasing its potential as a universal regularization method in deep learning.

To summarize, ChannelDropBack presents a forward-consistent approach to stochastic regularization and provides a universally applicable technique to enhance the performance and generalization of deep convolutional networks.

\section{Related Work}

Stochastic regularization techniques have become essential for deep learning, mitigating overfitting and enhancing generalization.  Dropout~\cite{srivastava2014dropout} randomly deactivates neurons during training, preventing complex co-adaptations and promoting the learning of robust features. DropConnect~\cite{wan2013regularization} generalizes dropout by randomly removing connections between neurons. However, both methods suffer from inconsistency between the trained and deployed networks. Stochastic Depth~\cite{huang2016deep} and DropPath~\cite{larsson2016fractalnet} address the challenge of training very deep networks by randomly dropping entire layers or paths during training. ScheduledDropPath~\cite{zoph2018learning} extends DropPath with a linearly decreasing drop rate. These methods, while effective, still maintain the discrepancy between the trained and deployed networks. Shake-Shake~\cite{gastaldi2017shake} and ShakeDrop~\cite{yamada2018shakedrop} regularization techniques apply random perturbations to the inputs of a network during training, encouraging the model to learn more robust features. However, they require modifications to the original architecture. DropBlock~\cite{ghiasi2018dropblock} and Spatial Dropout~\cite{tompson2015efficient} introduce stochasticity in a structured manner by dropping contiguous regions of feature maps or channels. Although these methods improve regularization, they still require the addition of specialized layers or are only effective for specific network topologies. In summary, existing stochastic training methods often necessitate architecture modifications, are limited to specific network topologies, and maintain a discrepancy between the trained and deployed networks.


\section{ChannelDropBack}

ChannelDropBack is a simple method for stochastic training approach for deep convolutional networks. It aims to improve regularization by introducing randomness exclusively into the backward information flow during training, while preserving the integrity of the forward pass, and ensuring that the same network is deployed during both training and inference phases. It allows for seamless integration into the training process of any model without the need to modify its architecture or add specialized layers.

\subsection{Layer and Channel Selection Strategy}

During each training iteration, ChannelDropBack randomly selects a single layer within the network for which weight updates will be applied to a subset of its channels or rows. The layer selection strategy is based on a predefined probability distribution, ensuring that each layer has a non-zero probability of being selected. In our experiments, we employ a uniform distribution for layer selection, but other distributions can also be explored.

Formally, let $L$ denote the set of all layers in the network, and let $l$ be a layer sampled from $L$ according to the predefined probability distribution. The selected layer is then given by:

\[ l_{\text{selected}} \in L \]

Once the layer, $l_{\text{selected}}$, has been selected, we randomly choose a subset of channels (in the case of convolutional layers) or rows (in the case of fully connected layers) to be updated during backpropagation. The number of channels or rows to be selected can be determined by a predefined hyperparameter, $p$, which represents the proportion of channels or rows to be updated. We always select from the first dimensions of the layer to drop.

Formally, let $C$ denote the set of all channels in a convolutional layer or the set of all rows in a fully connected layer. Let $c$ be a channel or row sampled from $C$ according to the predefined probability distribution. The subset of selected channels or rows, $S$, is then given by:

\[ S = \{c_1, c_2, \ldots, c_k\} \subseteq C \]

where $k$ is the number of selected channels or rows, and $c_1, c_2, \ldots, c_k$ are the randomly selected channels or rows.

\subsection{Backward Pass Modification}

Once the subset of channels or rows, $S$, has been selected, we modify the backpropagation process to apply weight updates only to the selected channels or rows in the selected layer, $l_{\text{selected}}$. During the backward pass, ChannelDropBack computes the gradients for all layers in the network as usual. However, when it comes to updating the weights, ChannelDropBack applies the updates only to the selected channels or rows in the selected layer, $l_{\text{selected}}$, while leaving the remaining channels or rows unchanged.

Formally, let $w_l$ denote the weights of layer $l$, and let $\Delta w_l$ denote the weight updates computed during backpropagation. The updated weights, $w'_l$, are given by:

\[ w'_l = w_l + \Delta w_l \cdot I(l = l_{\text{selected}}) \cdot I(c \in S) \]

where $I(l = l_{\text{selected}})$ is an indicator function that returns 1 if layer $l$ is the selected layer, $l_{\text{selected}}$, and 0 otherwise. $I(c \in S)$ is an indicator function that returns 1 if channel or row $c$ is in the subset $S$, and 0 otherwise.

\subsection{Training}

The training procedure with ChannelDropBack is similar to the standard training process, with the exception of the layer selection, channel or row selection, and backward pass modification steps. The overall training procedure can be summarized as follows:

\begin{algorithm}[H]
\caption{ChannelDropBack Training Procedure}
\begin{algorithmic}[1]
\State Initialize the network weights.
\For{each training iteration}
    \State Perform a forward pass to compute the output and loss.
    \State Randomly select a layer, $l_{\text{selected}}$, according to the layer selection strategy.
    \State Randomly select a subset of channels or rows, $S$, according to the channel or row selection strategy.
    \State Perform a backward pass to compute the gradients for all layers.
    \State Apply weight updates only to the selected channels or rows in the selected layer, $l_{\text{selected}}$, as described in the backward pass modification step.
\EndFor
\end{algorithmic}
\end{algorithm}

\subsubsection{Relation between Learning Rate and Drop Rate}

The drop rate in ChannelDropBack plays a crucial role in balancing the stochastic regularization and the learning process. To achieve optimal performance, we consider the relationship between the learning rate and the drop rate. We consider layer drop rate as the probability to select a layer for dropping, and channel drop rate as the probability to drop a channel within the selected layer. Here, we discuss three key aspects of this relationship and provide explanations for their underlying mechanisms.

\begin{enumerate}
\item \textbf{Starting with zero drop rate:} It is best to start with a zero layer drop rate until the first weight decaying occurs. The rationale behind this is to allow the model to learn the initial features and representations without being affected by the stochastic regularization introduced by ChannelDropBack. Starting with a zero layer drop rate enables the model to learn a good initial representation of the data, which can serve as a solid foundation for further refinement and optimization as the drop rate is gradually increased.

\item \textbf{Increasing layer drop rate and reducing learning rate:} After the each weight decaying, we increase the drop rate in ChannelDropBack. This recommendation stems from the observation that as training advances and weights are reduced, the model becomes more susceptible to overfitting. These results align with the findings presented by Morerio et al~\cite{morerio2017curriculum}.

\item \textbf{Layer skipping and drop probability:} Similar to \cite{he2016deep}, we observed that it is better to reduce stochasticity for the earlier layers for improved convergence of the model. We therefore employ a simple policy of skipping during the layer selection phase the initial four layers in the model, always training them regularly. We also investigated the survival probability approach of \cite{he2016deep} for our layer drop rate. However, as this method did not yield notable improvements, we opted to retain our simpler uniform probability approach.
\end{enumerate}
\subsection{Inference}

One of the key advantages of ChannelDropBack is that it ensures the exact same network is deployed during both training and inference. This is because ChannelDropBack does not introduce any modifications to the forward pass or the network architecture during training. Consequently, the inference procedure with ChannelDropBack is identical to the standard inference process, without any additional computations or modifications.

\section{Results}
\label{sec:results}
We evaluate ChannelDropBack on ImageNet \cite{deng2009imagenet}, CIFAR-100, CIFAR-10 \cite{krizhevsky2009learning}, and SVHN \cite{netzer2011reading} datasets using MobileNetv2 \cite{sandler2018mobilenetv2}, ShuffleNetV2 \cite{ma2018shufflenet}, EfficientNet-B0 \cite{tan2019efficientnet}, DenseNet121 \cite{iandola2014densenet}, ResNet-50 \cite{he2016deep}, and Vision Transformer (ViT-B) \cite{dosovitskiy2020image} architectures. We compare ChannelDropBack against baseline models and other regularization techniques. All experiments use SGD optimizer with momentum 0.9 and weight decay 1e-4, conducted on eight Tesla V100 SXM2 32Gb and eight Quadro RTX6000 GPUs.

\subsection{Transfer Learning Results}

We evaluate ChannelDropBack in transfer learning scenarios using pre-trained ViT models on ImageNet-21K and ResNet-50 on ImageNet-1K, fine-tuned on CIFAR10, CIFAR-100, and ImageNet-1K. CIFAR10 and CIFAR100 were fine-tuned for 85 epochs, and ImageNet for 10 epochs. We use a learning rate of 0.001 for ViT and 0.01 for ResNet50, with a batch size of 256.

Tables \ref{tab:transfer_learning} and \ref{tab:resnet_finetune} present the accuracy comparison between models fine-tuned with ChannelDropBack and those without stochastic regularization.

\begin{table}[tb]
\centering
\begin{tabular}{@{}lc|cc|cc|cc@{}}
\toprule
\multirow{2}{*}{Model} & \multirow{2}{*}{Resolution} & \multicolumn{2}{c|}{CIFAR10} & \multicolumn{2}{c|}{CIFAR100} & \multicolumn{2}{c}{ImageNet} \\
& & Ours & Baseline & Ours & Baseline & Ours & Baseline \\
\midrule
ViT-Base/16 & 224x224 & 98.80 & \textbf{98.95} & \textbf{92.20} & 91.67 & \textbf{82.03} & 81.32 \\
ViT-Large/16 & 224x224 & \textbf{99.08} & 99.04 & \textbf{93.44} & 93.25 & \textbf{83.13} & 82.94 \\
ViT-Base/16 & 384x384 & - & - & - & - & \textbf{83.90} & 83.32 \\
ViT-Large/16 & 384x384 & - & - & - & - & \textbf{85.05} & 85.05 \\
\bottomrule
\end{tabular}
\caption{Fine-tuned model accuracy on various datasets.}
\label{tab:transfer_learning}
\end{table}

\begin{table}[tb]
\centering
\begin{tabular}{@{}lccc@{}}
\toprule
Model & SVHN & CIFAR10 & CIFAR100 \\
\midrule
ResNet50 (Baseline) & 95.31 & 97.38 & 84.20 \\
ResNet50 (ChannelDropBack) & \textbf{95.37} & \textbf{97.45} & \textbf{84.90} \\
\bottomrule
\end{tabular}
\caption{Fine-tuned ResNet50 accuracy on various datasets. The model was pre-trained on ImageNet-1K. Pre-train and fine-tune images were resized to 224x224.}
\label{tab:resnet_finetune}
\end{table}

ChannelDropBack enhances transfer learning capabilities by introducing controlled stochasticity during the fine-tuning process. This stochasticity helps prevent overfitting to the source domain, allowing the model to adapt more effectively to the target domain. The selective channel dropping encourages the network to learn robust features that generalize well across datasets.

\subsection{Full Training Results}

We evaluate training from scratch using ChannelDropBack with a multi-step linear learning rate schedule. The initial learning rate is 0.01, with milestones at 0.3, 0.6, and 0.8 of the total 200 epochs. The ChannelDropBack drop rate starts at 0.01 and increases linearly to 0.3 after the last learning rate decay. We exclude the first layers from dropping to improve the learning process.

Table \ref{tab:pref_full_train} presents the top-1 accuracy on CIFAR-100 and CIFAR-10 test sets for different models.

\begin{table}[tb]
\centering
\begin{tabular}{@{}lcc|cc@{}}
\toprule
\multirow{2}{*}{Model} &  \multicolumn{2}{c|}{CIFAR10} & \multicolumn{2}{c}{CIFAR100} \\
& ChannelDropBack & Baseline & ChannelDropBack & Baseline \\
\midrule
MobileNetv2 &\textbf{94.41} & 94.04 & \textbf{73.75} & 68.08 \\
EfficientNetB0 & \textbf{93.16} & 92.79 & \textbf{67.23} & 67.15 \\
ShuffleNetV2  & \textbf{93.13} & 92.65 & \textbf{65.45} & 65.32 \\
DenseNet121  & \textbf{95.52} & 95.12 & \textbf{77.30} & 77.01 \\
ResNet50 &89.12 & \textbf{89.16} & \textbf{77.55} & 76.51 \\
\bottomrule
\end{tabular}
\caption{Model Accuracy on various datasets trained from scratch.}
\label{tab:pref_full_train}
\end{table}

ChannelDropBack consistently outperforms the baseline model across datasets and architectures, demonstrating its universal applicability.

\subsection{Comparison with Other Regularization Techniques}

We compare ChannelDropBack with Dropout, SpatialDropout \cite{Tompson_2015_CVPR}, and Dropblock \cite{ghiasi2018dropblock}. Table \ref{tab:comparison} summarizes the top-1 accuracy for ResNet-50 on CIFAR-100. As can be seen, ChannelDropBack outperforms other regularization techniques.

\begin{table}[h]
\centering
\begin{tabular}{@{}lc@{}}
\toprule
Method & Top-1 Accuracy (\%) \\
\midrule
Baseline & 76.51 \\
Dropout (kp=0.7) \cite{srivastava2014dropout} & 76.80 \\
DropPath (kp=0.9) \cite{larsson2016fractalnet} & 77.10 \\
SpatialDropout (kp=0.9) \cite{tompson2015efficient} & 77.41 \\
Dropblock (kp=0.9) \cite{ghiasi2018dropblock} & 77.42 \\
ChannelDropBack & \textbf{77.55} \\
\bottomrule
\end{tabular}
\caption{Comparison of regularization techniques on CIFAR-100 with ResNet-50.}
\label{tab:comparison}
\end{table}

\subsection{Performance on Different Network Depths}

We investigate ChannelDropBack performance across varying network depths using ResNet architectures. Table \ref{tab:network_depth} presents results for ResNet-18, ResNet-34, ResNet-50, and ResNet-101 on CIFAR-100. ChannelDropBack consistently improves performance across network depths, demonstrating effectiveness for both shallow and deep networks.

\begin{table}[ht]
\centering
\begin{tabular}{@{}lcc@{}}
\toprule
Model & ChannelDropBack & Baseline \\
\midrule
ResNet18 & \textbf{75.08} (+0.35) & 74.73 \\
ResNet34 & \textbf{76.28} (+0.72) & 75.56 \\
ResNet50 & \textbf{77.55} (+1.04) & 76.51 \\
ResNet101 & \textbf{77.67} (+0.86) & 76.81 \\
\bottomrule
\end{tabular}
\caption{ChannelDropBack performance on ResNet architectures of different depths for CIFAR-100. Values in parentheses indicate improvement over the baseline.}
\label{tab:network_depth}
\end{table}

\subsection{Layer Drop Rate Policy}

We analyze the impact of different layer drop rate policies on ChannelDropBack. Table \ref{tab:ablation} shows results for ResNet-34 on CIFAR-100. Fixed drop rate applies channel dropping operations from the first epoch without incremental increase. Adaptive is as discussed in Section 3.3. "without skipping first layers" allows also dropping of the first layers.

\begin{table}[tb]
\centering
\begin{tabular}{@{}lc@{}}
\toprule
Configuration & Top-1 Accuracy (\%) \\
\midrule
Baseline & 75.56 \\
ChannelDropBack  &  75.88 \\
ChannelDropBack (Adaptive) & \textbf{76.28} \\
ChannelDropBack (Adaptive without skipping first layers) & 75.73 \\
\bottomrule
\end{tabular}
\caption{Study results for different drop rate policies for ResNet-34 on CIFAR-100.}
\label{tab:ablation}
\end{table}

Both the adaptive drop rate strategy and skipping layer selection mechanism contribute to ChannelDropBack's effectiveness. Using a fixed drop rate throughout training and remove skipping results in a decrease in accuracy.

\subsection{Impact of Channel Drop Rate}

\begin{figure}[h]
\centering
\includegraphics[width=0.8\textwidth]{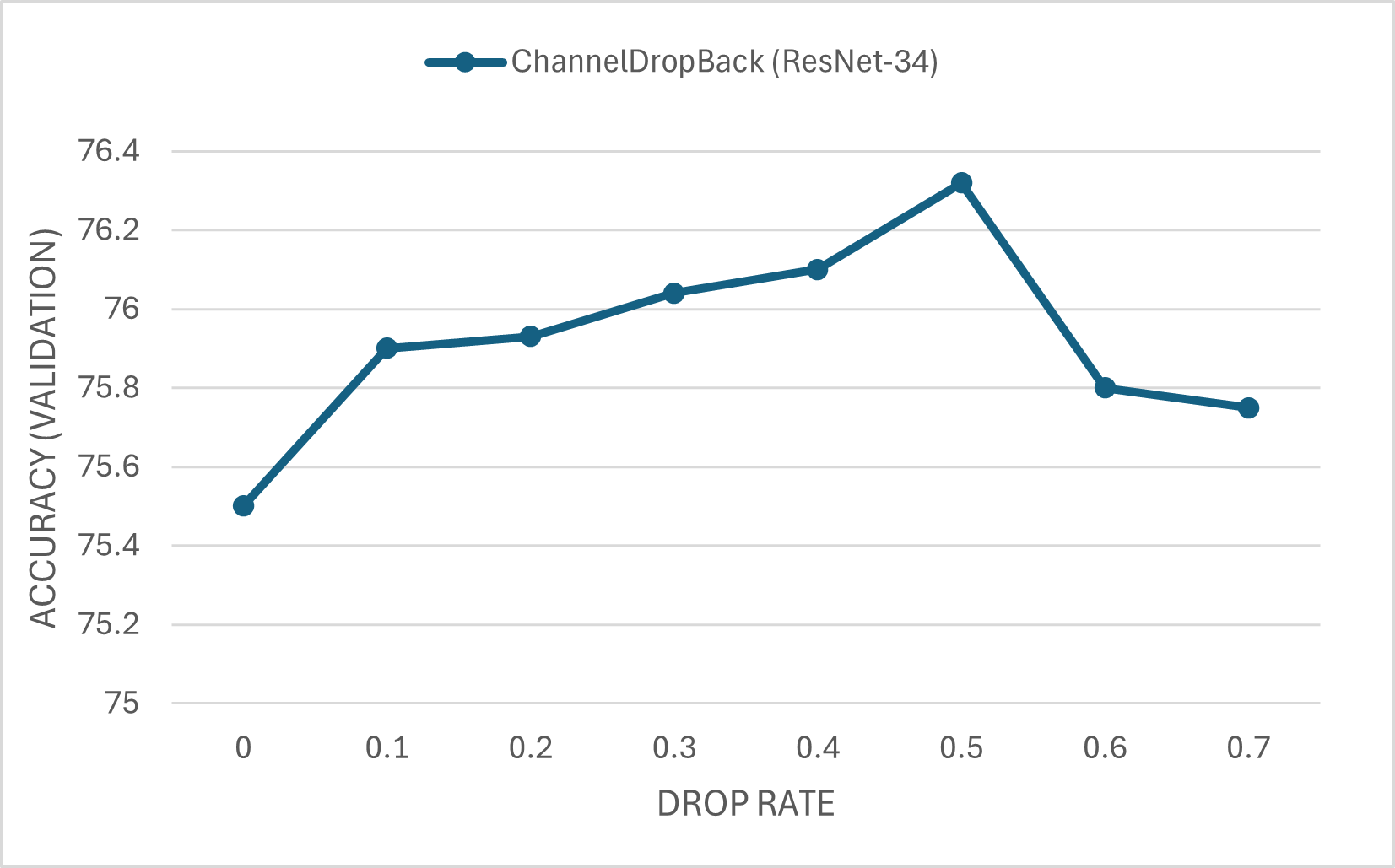}
\caption{Impact of the channel drop rate on top-1 accuracy for ResNet-34 on CIFAR-100.}
\label{fig:drop_rate_impact}
\end{figure}

Figure \ref{fig:drop_rate_impact} illustrates the relationship between the drop rate of the channels and top-1 accuracy on CIFAR-100 using ResNet-34.

Accuracy improves with increasing drop rate up to an optimal point, after which performance declines. This suggests an optimal level of randomness that maximizes performance. For CIFAR-100 with ResNet-34, a drop rate of 0.5 yielded the best results.

\subsection{Computational Overhead}

We assessed the scalability and computational overhead of ChannelDropBack. Table \ref{tab:overhead} compares the training time per epoch for ResNet-50 on CIFAR-100 with and without ChannelDropBack. ChannelDropBack incurs negligible computational overhead, due to the balance between reduced backpropagation operations and additional selection overhead.

\begin{table}[tb]
\centering
\begin{tabular}{@{}lc@{}}
\toprule
Configuration & Time (seconds) \\
\midrule
Baseline & 29.1 \\
ChannelDropBack & 29.3 \\
\bottomrule
\end{tabular}
\caption{Training time per epoch for ResNet-50 on CIFAR-100, comparing baseline and ChannelDropBack configurations.}
\label{tab:overhead}
\end{table}





\section{Conclusion}

ChannelDropBack is a simple yet effective stochastic training approach that improves regularization in deep networks by introducing randomness exclusively into the backward information flow during training. By randomly selecting a subset of channels or rows for weight updates in a selected layer and preserving the integrity of the forward pass, ChannelDropBack offers a universally applicable regularization technique, without requiring modifications to the network architecture. This makes it easy to implement across a wide range of network topologies, from traditional convolutional networks like ResNet and EfficientNet to more recent architectures like Vision Transformers (ViT). By maintaining consistency between the training and inference phases, ChannelDropBack also ensures that the trained model is identical to the deployed model, eliminating the potential discrepancies that can arise with other stochastic regularization methods.

\subsubsection{Acknowledgements} We acknowledge the Ariel HPC Center at Ariel University for providing computing resources that have contributed to the research results reported within this paper.

\bibliographystyle{splncs04}
\bibliography{references}

\begin{thebibliography}{10}
\providecommand{\url}[1]{\texttt{#1}}
\providecommand{\urlprefix}{URL }
\providecommand{\doi}[1]{https://doi.org/#1}

\bibitem{10095537}
Aharon, S., Ben-Artzi, G.: Hypernetwork-based adaptive image restoration. In: ICASSP 2023 - 2023 IEEE International Conference on Acoustics, Speech and Signal Processing (ICASSP). pp.~1--5 (2023). \doi{10.1109/ICASSP49357.2023.10095537}

\bibitem{deng2009imagenet}
Deng, J., Dong, W., Socher, R., Li, L.J., Li, K., Fei-Fei, L.: Imagenet: A large-scale hierarchical image database. In: 2009 IEEE conference on computer vision and pattern recognition. pp. 248--255. Ieee (2009)

\bibitem{dosovitskiy2020image}
Dosovitskiy, A., Beyer, L., Kolesnikov, A., Weissenborn, D., Zhai, X., Unterthiner, T., Dehghani, M., Minderer, M., Heigold, G., Gelly, S., et~al.: An image is worth 16x16 words: Transformers for image recognition at scale. arXiv preprint arXiv:2010.11929  (2020)

\bibitem{gastaldi2017shake}
Gastaldi, X.: Shake-shake regularization. arXiv preprint arXiv:1705.07485  (2017)

\bibitem{ghiasi2018dropblock}
Ghiasi, G., Lin, T.Y., Le, Q.V.: Dropblock: A regularization method for convolutional networks. Advances in neural information processing systems  \textbf{31} (2018)

\bibitem{he2016deep}
He, K., Zhang, X., Ren, S., Sun, J.: Deep residual learning for image recognition. In: Proceedings of the IEEE conference on computer vision and pattern recognition. pp. 770--778 (2016)

\bibitem{9380520}
Hel-Or, Y., Ben-Artzi, G.: The role of redundant bases and shrinkage functions in image denoising. IEEE Transactions on Image Processing  \textbf{30},  3778--3792 (2021). \doi{10.1109/TIP.2021.3065226}

\bibitem{huang2016deep}
Huang, G., Sun, Y., Liu, Z., Sedra, D., Weinberger, K.Q.: Deep networks with stochastic depth. In: Computer Vision--ECCV 2016: 14th European Conference, Amsterdam, The Netherlands, October 11--14, 2016, Proceedings, Part IV 14. pp. 646--661. Springer (2016)

\bibitem{iandola2014densenet}
Iandola, F., Moskewicz, M., Karayev, S., Girshick, R., Darrell, T., Keutzer, K.: Densenet: Implementing efficient convnet descriptor pyramids. arXiv preprint arXiv:1404.1869  (2014)

\bibitem{ioffe2015batch}
Ioffe, S., Szegedy, C.: Batch normalization: Accelerating deep network training by reducing internal covariate shift. In: International conference on machine learning. pp. 448--456. pmlr (2015)

\bibitem{kingma2014adam}
Kingma, D.P., Ba, J.: Adam: A method for stochastic optimization. arXiv preprint arXiv:1412.6980  (2014)

\bibitem{krizhevsky2009learning}
Krizhevsky, A., Hinton, G., et~al.: Learning multiple layers of features from tiny images  (2009)

\bibitem{larsson2016fractalnet}
Larsson, G., Maire, M., Shakhnarovich, G.: Fractalnet: Ultra-deep neural networks without residuals. arXiv preprint arXiv:1605.07648  (2016)

\bibitem{ma2018shufflenet}
Ma, N., Zhang, X., Zheng, H.T., Sun, J.: Shufflenet v2: Practical guidelines for efficient cnn architecture design. In: Proceedings of the European conference on computer vision (ECCV). pp. 116--131 (2018)

\bibitem{menghani2023}
Menghani, G.: Efficient deep learning: A survey on making deep learning models smaller, faster, and better. ACM Computing Surveys  \textbf{55}(12),  1--37 (2023)

\bibitem{morerio2017curriculum}
Morerio, P., Cavazza, J., Volpi, R., Vidal, R., Murino, V.: Curriculum dropout. In: Proceedings of the IEEE International Conference on Computer Vision. pp. 3544--3552 (2017)

\bibitem{netzer2011reading}
Netzer, Y., Wang, T., Coates, A., Bissacco, A., Wu, B., Ng, A.Y., et~al.: Reading digits in natural images with unsupervised feature learning. In: NIPS workshop on deep learning and unsupervised feature learning. vol.~2011, p.~4. Granada (2011)

\bibitem{ofir2022smm}
Ofir, A., Ben-Artzi, G.: Smm-conv: Scalar matrix multiplication with zero packing for accelerated convolution. In: Proceedings of the IEEE/CVF Conference on Computer Vision and Pattern Recognition. pp. 3067--3075 (2022)

\bibitem{sandler2018mobilenetv2}
Sandler, M., Howard, A., Zhu, M., Zhmoginov, A., Chen, L.C.: Mobilenetv2: Inverted residuals and linear bottlenecks. In: Proceedings of the IEEE conference on computer vision and pattern recognition. pp. 4510--4520 (2018)

\bibitem{srivastava2014dropout}
Srivastava, N., Hinton, G., Krizhevsky, A., Sutskever, I., Salakhutdinov, R.: Dropout: a simple way to prevent neural networks from overfitting. The journal of machine learning research  \textbf{15}(1),  1929--1958 (2014)

\bibitem{tan2019efficientnet}
Tan, M., Le, Q.: Efficientnet: Rethinking model scaling for convolutional neural networks. In: International conference on machine learning. pp. 6105--6114. PMLR (2019)

\bibitem{tompson2015efficient}
Tompson, J., Goroshin, R., Jain, A., LeCun, Y., Bregler, C.: Efficient object localization using convolutional networks. In: Proceedings of the IEEE conference on computer vision and pattern recognition. pp. 648--656 (2015)

\bibitem{Tompson_2015_CVPR}
Tompson, J., Goroshin, R., Jain, A., LeCun, Y., Bregler, C.: Efficient object localization using convolutional networks. In: Proceedings of the IEEE Conference on Computer Vision and Pattern Recognition (CVPR) (June 2015)

\bibitem{wan2013regularization}
Wan, L., Zeiler, M., Zhang, S., Le~Cun, Y., Fergus, R.: Regularization of neural networks using dropconnect. In: International conference on machine learning. pp. 1058--1066. PMLR (2013)

\bibitem{yamada2018shakedrop}
Yamada, Y., Iwamura, M., Kise, K.: Shakedrop regularization  (2018)

\bibitem{zoph2018learning}
Zoph, B., Vasudevan, V., Shlens, J., Le, Q.V.: Learning transferable architectures for scalable image recognition. In: Proceedings of the IEEE conference on computer vision and pattern recognition. pp. 8697--8710 (2018)

\end{thebibliography}
\end{document}